\documentclass[10pt,twocolumn,letterpaper]{article}

\usepackage{cvpr}              

\usepackage{graphicx}
\usepackage{amsmath}
\usepackage{amssymb}
\usepackage{booktabs}
\usepackage{bm}
\usepackage{multirow}

\usepackage[pagebackref,breaklinks,colorlinks]{hyperref}

\usepackage[capitalize]{cleveref}
\crefname{section}{Sec.}{Secs.}
\Crefname{section}{Section}{Sections}
\Crefname{table}{Table}{Tables}
\crefname{table}{Tab.}{Tabs.}

\def\cvprPaperID{2990} 
\def\confName{CVPR}
\def\confYear{2023}

\begin{document}

\title{Few-Shot Learning with Visual Distribution Calibration and Cross-Modal Distribution Alignment}

\author{
Runqi Wang{$^{1,2}$}{$^*$}, Hao Zheng{$^{2,3}$}{$^*$}, Xiaoyue Duan{$^1$}\thanks {Co-first author.},
Jianzhuang Liu{$^{2}$},\\ Yuning Lu{$^{2,4}$},
Tian Wang{$^{1}$}, Songcen Xu{$^{2}$}, Baochang Zhang{$^{1,5}$}\thanks {Corresponding author.} \\ 
{$^1$}Beihang University \ {$^2$}Huawei Noah's Ark Lab \ {$^3$}Tokyo Institute of Technology \\ {$^4$}University of Science and Technology of China \ {$^{5}$}Zhongguancun Laboratory
}

\maketitle

\begin{abstract}

Pre-trained vision-language models have inspired much research on few-shot learning. However, with only a few training images, there exist two crucial problems: (1) the {visual feature distributions are easily distracted by class-irrelevant information in images}, and (2) the alignment between the visual and language feature distributions is difficult. To deal with the distraction problem, we propose a Selective Attack module, which consists of trainable adapters that generate spatial attention maps of images to guide the attacks on {class-irrelevant image areas.}
{By messing up these areas,} the critical features are captured and the {visual distributions of image features are calibrated.} To better align the visual and language feature distributions that describe the same object class, we propose a cross-modal distribution alignment module, in which we introduce a vision-language prototype for each class to align the distributions, and adopt the Earth Mover's Distance (EMD) to optimize the prototypes. For efficient computation, the upper bound of EMD is derived. In addition, we propose an augmentation strategy to increase the diversity of the images and the text prompts, which can reduce overfitting to the few-shot training images. Extensive experiments on 11 datasets demonstrate that our method consistently outperforms prior arts in few-shot learning. The implementation code will be available at \href{https://github.com/bhrqw/SADA}{https://github.com/bhrqw/SADA}.

\end{abstract}

\vspace{-2mm}
\section{Introduction}
\label{sec:introduction}

Thanks to the availability of large-scale datasets and well-designed training strategies, the performances of many computer vision tasks have been greatly improved. Recent progress in vision-language models (VLMs), such as CLIP~\cite{radford2021learning} and ALIGN~\cite{jia2021scaling}, provides a promising way towards utilizing human language to address downstream recognition tasks efficiently. 
As vision and language usually contain complementary information, joint learning of image and text representations has proven quite effective. 
{
Although CLIP has demonstrated impressive zero-shot learning capability, it is still challenging to better adapt it to downstream tasks. Naively fine-tuning CLIP on downstream datasets has limited effect, since it may destroy the prior learned from the massive data during pre-training. Therefore, effective transfer methods are needed to boost the downstream performances of CLIP.}
In order to maintain the capability of pre-trained VLMs and further boost downstream performances, different approaches have been proposed to fine-tune a small proportion of additional parameters while keeping the pre-trained parameters frozen. Among these approaches, prompt learning~\cite{zhou2022learning,zhou2022conditional} and visual adapters~\cite{zhang2021tip,gao2021clip} are two common approaches. However, the lack of training samples in few-shot settings increases the risk of overfitting the trained prompts or adapters. The class-irrelevant features (\emph{e.g.}, the cluttered image backgrounds) drive the image features far away from their true distributions of the same category. 
Besides, VLMs such as CLIP have such a problem that the distributions of the image and text features are not really aligned~\cite{ramesh2022hierarchical}, and the problem becomes more challenging in few-shot settings. {Therefore, the visual distributions should be calibrated by reducing class-irrelevant image contents, and the distributions of image and text features should be further aligned,} so as to promote the model's learning of class-relevant critical features. The purpose of this paper is to develop an effective VLM transfer strategy for few-shot learning to solve the above problems with \textit{Selective Attack} (SA) and \textit{Cross-Modal Distribution Alignment} (CMDA).

\begin{figure*}[ht]
    \begin{center}
    \includegraphics[width=0.75\linewidth]{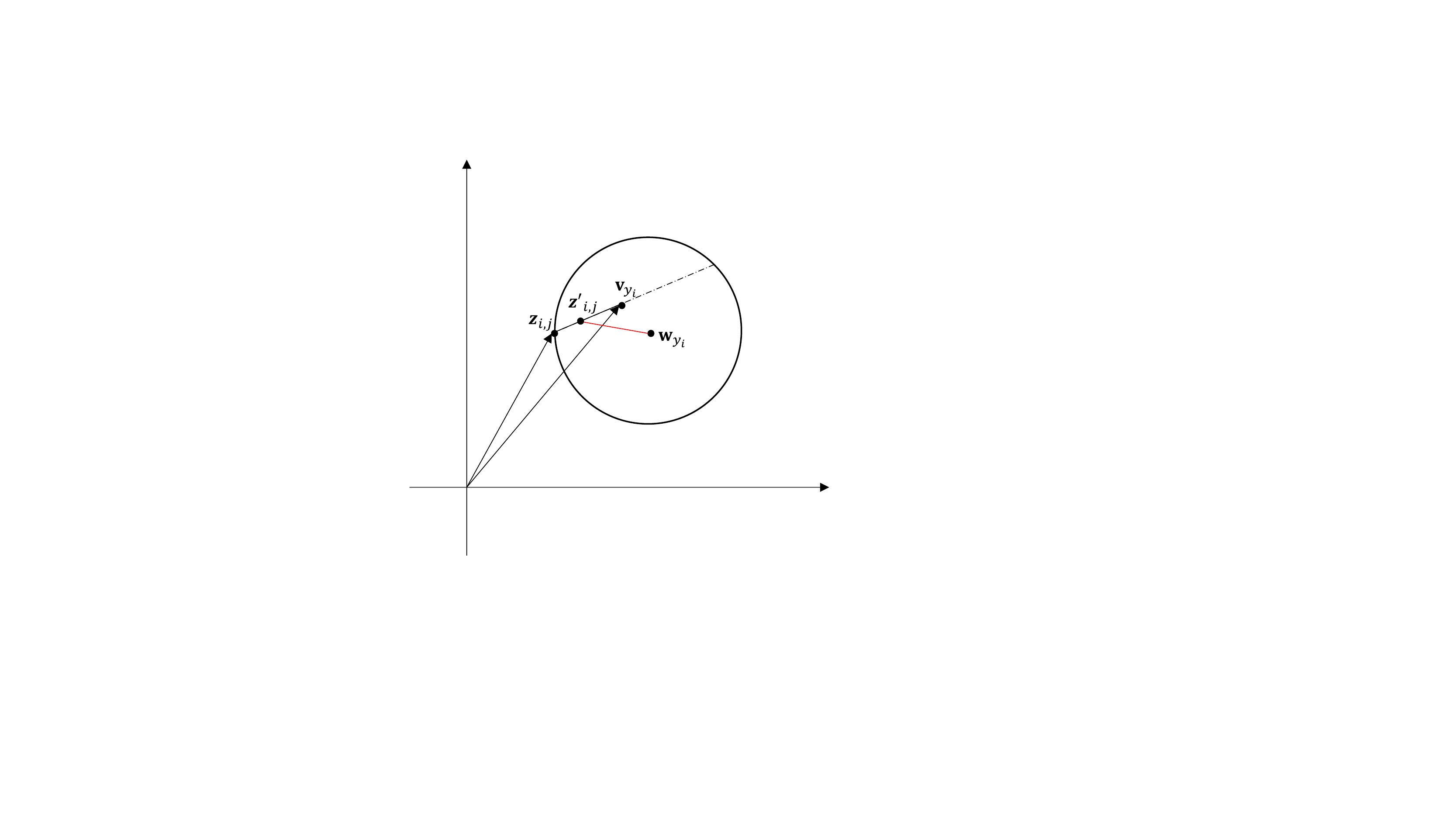}
    \vspace{-2mm}
    \caption{(a) The t-SNE~\cite{van2008visualizing} visualization of the image feature distribution before Selective Attack, where the features are obtained by the CLIP image encoder on the CIFAR10 dataset. The dots in different colors represent different classes of the image features. (b) After Selective Attack, the intra-class distribution is significantly more compact. (c) The distribution histograms of image features and text features of the same class (`bird') on CIFAR10 before CMDA, {where the horizontal axis denotes the value of each element of the feature vectors, and the vertical axis denotes the number of elements.
    } (d) After CMDA, the difference between the two distributions is significantly reduced.}
    \vspace{-7mm}
    \label{fig:motivation}
\end{center}
\end{figure*}

Images often contain class-irrelevant information, which is also embedded into the image representations. With only a few samples, the model can easily learn these cluttered representations, resulting in overfitting. This seriously hinders the learning of critical features that help the model recognize unseen samples. To solve this problem, we propose the SA module, which consists of two trainable adapters that generate a kernelized attention map to locate the class-irrelevant areas of the images. {The attention is adopted to guide Gaussian perturbations to attack images before they are fed into the image encoder. By messing up these class-irrelevant image contents through SA, 
}
{we facilitate the model's learning of truly critical features that can be transferred to recognize new samples within the same category.}
As an example in Figs.~\ref{fig:motivation} (a) and (b), after Selective Attack (SA), the distributions of the image features are calibrated, and the intra-class features become obviously more clustered.

\vspace{-2mm}
Another challenge is that the distributions of the image and the text representations of the same class are not truly aligned in CLIP~\cite{ramesh2022hierarchical} as shown in Fig.~\ref{fig:motivation}(c). The unaligned distributions lead to inaccurate similarity calculations between image features and text features during inference, resulting in incorrect predictions. The lack of samples in few-shot settings further makes the problem even more serious. To address it, we propose a CMDA module, in which we construct a Vision-Language Prototype (VLP) for each class to promote the cross-modal distribution alignment. Specifically, the element values of VLP are initialized by averaging all the image representations from the corresponding class. During training, each VLP is optimized by reducing its distance to the language prototype (defined in Sec.~\ref{sec:3.4}) of the same class, thus promoting the cross-modal distribution alignment. The Earth Mover's Distance (EMD) is a suitable metric for the alignment, which can not only reflect the similarity between two distributions but also represent the minimal transmission cost~\cite{zhang2020deepemd}. We derive a concise upper bound of the EMD distance, which can balance the performance and computational consumption. As shown in Figs.~\ref{fig:motivation} (c) and (d), the effect of Cross-Modal Distribution Alignment (CMDA) is obvious that the difference between the image and text feature distributions is effectively reduced. In this way, the image features after CMDA can be better predicted by the text features.

\begin{figure*}[ht]
    \begin{center}
    \includegraphics[width=0.75\linewidth]{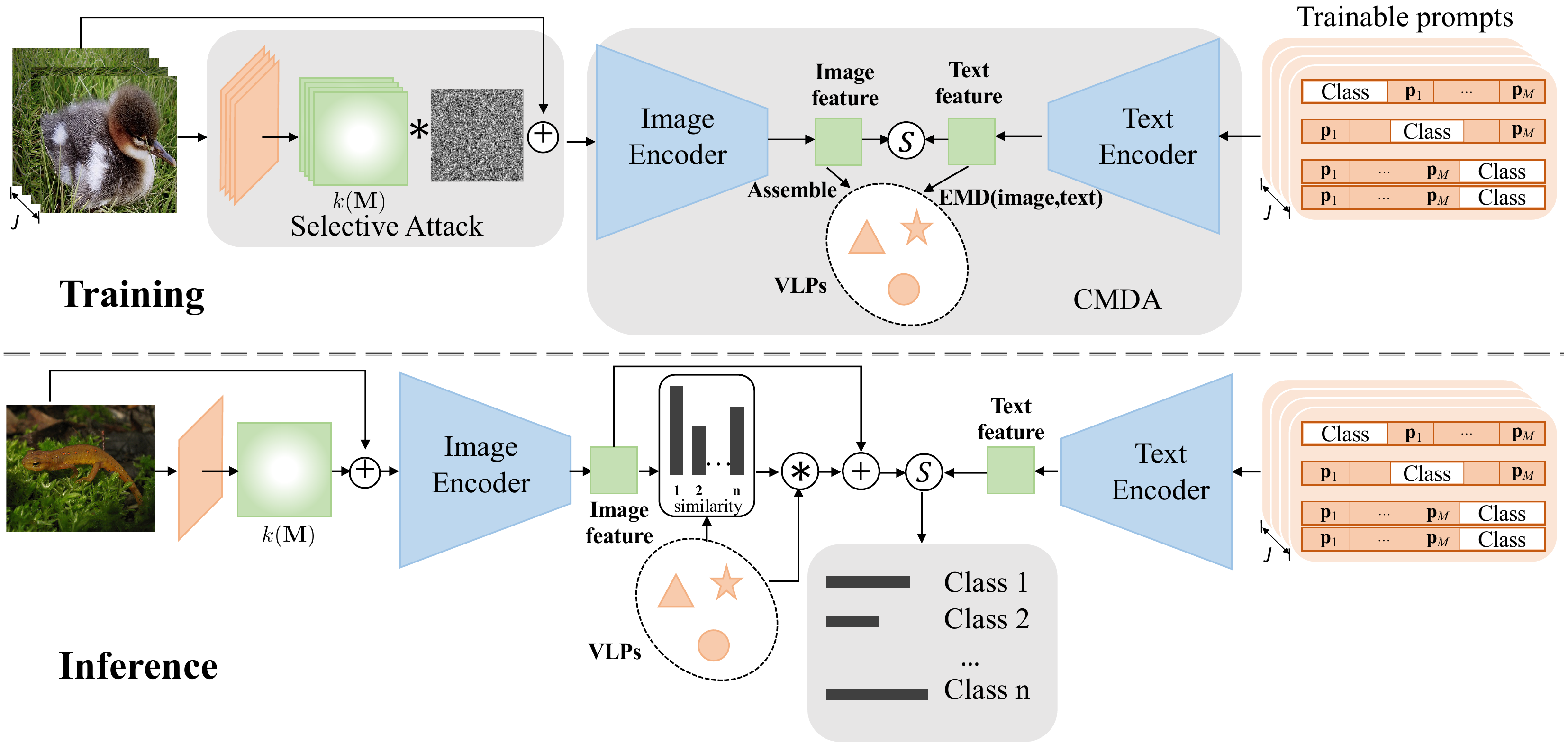}
    \vspace{-2mm}
    \caption{Overview of our framework. We introduce a \textit{Selective Attack} module to reduce the intra-class distances of image features during training. We also design a \textit{Cross-Modal Distribution Alignment} (CMDA) module to align the distributions of image and text representations.
    During training, the trainable parameters are denoted in orange and the encoders of CLIP are frozen. $J$: the number of augmentations; $\circledS$: cosine similarity computation; $\circledast$: element-wise product.}
    \vspace{-7mm}
    \label{fig:frame}
\end{center}
\end{figure*}

Automatic prompt learning for pre-trained VLMs has been proposed to reduce the expensive cost of hand-crafted prompt engineering~\cite{zhou2022learning}. However, the learned prompts may suffer from more overfitting than manual prompts~\cite{zhou2022conditional}. Therefore, instead of learning one soft prompt, we learn a distribution over a collection of prompts, as in ProDA~\cite{lu2022prompt}. Moreover, we introduce an augmentation strategy to increase the diversity of the images and the prompts. Specifically, we search for the four best augmentations from a collection of predefined ones. Using these operations, each image is augmented into four different forms. The collection of prompts is also divided into four groups, with each group trained by images in the corresponding augmentation form. Through the strategy, we improve the diversity of the images and the prompts, and fully excavate the semantic information in the prompts. The framework of our method is shown in Fig.~\ref{fig:frame}. Our contributions are summarized as follows:

\begin{itemize}
\vspace{-2mm}
\item 
We conduct Selective Attack on the class-irrelevant regions of images with the guidance of the attention generated by two trainable adapters to facilitate the model's learning of class-related features, which {calibrates the visual distributions.}

\vspace{-2mm}
\item
We propose Cross-Modal Distribution Alignment optimized by an EMD loss. The upper bound of EMD for Gaussian distribution is further derived for computation efficiency.

\vspace{-2mm}
\item
We present an augmentation strategy to reduce overfitting and increase the diversity of images and prompts. 

\vspace{-2mm}
\item
Our method outperforms prior arts in few-shot learning on 11 benchmarks.

\end{itemize}

\vspace{-5mm}
\section{Related Work}
\vspace{-1mm}
\subsection{Vision-Language Models}
\vspace{-1mm}
Recently, many vision-language models have demonstrated great potential in learning generic visual representations such as CLIP~\cite{radford2021learning}, ALIGN~\cite{jia2021scaling} and Flamingo~\cite{alayrac2022flamingo}. Learning under a large number of images and their text descriptions, VLMs are robust to distribution shifts and thus can transfer across different domains. CLIP adopts a two-stream architecture, consisting of an image encoder and a text encoder that encode image and text inputs separately and produce individual vision and language representations embedded in a joint space using a contrastive loss. The success of VLMs has inspired research on a series of downstream tasks such as image classification~\cite{radford2021learning}, object detection~\cite{du2022learning, gu2021open}, semantic segmentation~\cite{xu2021simple}, action recognition~\cite{wang2021actionclip}, video caption~\cite{tang2021clip4caption} and so on.

\vspace{-1mm}
\subsection{Few-Shot Learning}
\vspace{-1mm}
As a challenging problem, few-shot learning aims to adapt a model to a new task with just a few examples. Researchers explore meta-learning to find well-initialized models suitable for adaptation~\cite{li2017meta, bertinetto2018meta, zintgraf2019fast}, or compensate for the data insufficiency in few-shot settings by data augmentation~\cite{antoniou2019assume, qin2020diversity}. Other approaches improve few-shot accuracy through feature calibration. For example, MatchingNet~\cite{vinyals2016matching} and ProtoNet~\cite{snell2017prototypical} learn to classify samples by comparing their distances to the prototypes, \emph{i.e.}, the representatives of classes, while other approaches attempt to augment feature representations by leveraging intra-class variance \cite{liu2020deep, park2020meta}. Recently, VLMs are also used for few-shot learning. CLIP-Adapter~\cite{gao2021clip} adds an adapter after the CLIP image encoder, and finetunes it while freezing the encoders of CLIP. CoOp~\cite{zhou2022learning} turns a prompt into a set of continuous vectors which can be optimized end-to-end with the help of a few labeled data from the target dataset.
CoCoOp \cite{zhou2022conditional} is further proposed based on CoOp to learn dynamic prompts for each instance, boosting the generalization of prompts to unseen classes or datasets. However, continuous prompts suffer from more serious overfitting than manual prompts~\cite{zhou2022conditional}. Therefore, ProDA~\cite{lu2022prompt} proposes to learn a distribution over a collection of prompts instead of only one prompt. 

Differently, we introduce Selective Attack on class-irrelevant contents to facilitate the learning of transferable class-relevant features, and propose an augmentation strategy to increase the diversity of images and prompts, better alleviating overfitting. By constructing and optimizing the VLPs, our method aligns the cross-modal distributions of image and text features, thus achieving better few-shot accuracy.

\vspace{-2.5mm}
\section{Methodology}
\vspace{-1.5mm}
In this section, we first revisit prompt learning in Sec.~\ref{sec:3.1}, and present our augmentation strategy to increase the diversity of the images and the prompts in Sec.~\ref{sec:3.2}. Then, we propose our \textit{Selective Attack} (SA) module and \textit{Cross-Modal Distribution Alignment} (CMDA) module in Secs.~\ref{sec:3.3} and~\ref{sec:3.4} respectively. The overview of our framework is given in Fig.~\ref{fig:frame}.

\vspace{-1.5mm}
\subsection{Prompt Learning}
\label{sec:3.1}
\vspace{-1mm}

CLIP consists of an image encoder $f(\cdot)$ and a text encoder $g(\cdot)$. Specifically, the image $\mathbf{x}$ and the text $\mathbf{t}$ are fed into $f(\cdot)$ and $g(\cdot)$ respectively to obtain the image feature $\mathbf{z}\in\mathbb{R}^{D}$ and the text feature $\mathbf{w}\in\mathbb{R}^{D}$, where $\mathbf{t}$ is the input embedding which is obtained by feeding the raw text through an embedding layer. In CLIP, $\mathbf{t}$ is obtained via one of the hand-crafted prompts which have a template like ``a photo of a [CLS]", where [CLS] is a class name of the downstream task. Thus, the probability of predicting the testing image $\mathbf{x}_i$ as the class $y_i$ can be computed by:

\begin{equation}
    p(y_i|\mathbf{x}_i)=\frac{e^{\langle\mathbf{z}_i,\mathbf{w}_{y_i}\rangle/\tau}}{\sum_{k=1}^K e^{\langle\mathbf{z}_i,\mathbf{w}_k\rangle/\tau}},
    \label{eq:img_text_sim}
\end{equation}
where $\tau$ is a temperature parameter learned by CLIP, $\langle\cdot,\cdot\rangle$ denotes cosine similarity, $\mathbf{w}_k$ is derived from the text description $\mathbf{t}_k$ of the $k$-th class, and $K$ is the total number of downstream dataset classes. 

To bring about improvement in few-shot learning, methods have been proposed to fine-tune a small proportion of newly introduced parameters while keeping the CLIP encoders frozen. Among them, prompt learning achieves
{impressive}
performance. A representative of prompt learning is CoOp~\cite{zhou2022learning}, which learns a continuous prompt $\mathbf{P}$ instead of adopting hand-crafted prompt templates. Specifically, by concatenating $\mathbf{P}$ with the embedding of 
a class name, the text description $\mathbf{t}_k(\mathbf{P})$ of the $k$-th class is obtained as:
\begin{equation}
    \mathbf{t}_k(\mathbf{P}) = [\mathbf{p}]_1[\mathbf{p}]_2\dots[\mathbf{p}]_M[\mathbf{CLS}]_k,
    \label{eq:coop_prompt}
\end{equation}
where each $[\mathbf{p}]_m$, $m\in \{1,\dots,M\}$, is a learnable vector of $\mathbf{P}$ with the same dimension as the embedding of $[\mathbf{CLS}]_k$, and $\mathbf{P}$ is shared among all classes, $[\mathbf{CLS}]_k$ is the text embedding of the $k$-th class name, which can also appear at the start and middle of the prompt in our method. In this way, $\mathbf{w}_k$ in Eq.~\ref{eq:img_text_sim} is replaced by $g(\mathbf{t}_k(\mathbf{P}))$. By minimizing the difference between the outputs of the image and text encoders, the prompt can be optimized to facilitate the learning of class-relevant object contents. The objective function of prompt learning is thus obtained as:
\begin{equation}
    \mathcal{L}(\mathbf{P})=\mathbb{E}[-\text{log}\frac{e^{\langle\mathbf{z}_i,g(\mathbf{t}_{y_i}(\mathbf{P}))\rangle/\tau}}{\sum_{k=1}^K e^{\langle\mathbf{z}_i,g(\mathbf{t}_k(\mathbf{P}))\rangle/\tau}}].
    \label{eq:loss_coop}
\end{equation}

Prompt learning suffers from serious overfitting, as mentioned in~\cite{zhou2022conditional}. Therefore, we adopt the prompt learning strategy proposed in~\cite{lu2022prompt} to learn a distribution over diverse prompts instead of one single prompt, so as to capture the variance of visual representations. To further overcome overfitting, we additionally introduce an augmentation strategy to increase the diversity of the images and the prompts, as described in Sec.~\ref{sec:3.2}.

\subsection{Augmentation Strategy} 
\label{sec:3.2}

Augmentation is an intuitive way to increase data diversity. In our strategy, we set up a pool of common candidate augmentation operations, which contains operations: \textit{rotating, flipping, random gray scaling, random cropping$+$resizing, resizing, color jittering}, and \textit{Gaussian blurring}. We finally choose $J$ operations with the best results as our augmentation set during training. These $J$ operations are applied to each training image $\mathbf{x}_i$ to obtain $J$ augmented images $\mathbf{x}_{i,j},~j\in\{1,\dots,J\}$. In addition to augmenting images, the text prompts in CLIP should also be diverse enough to prevent overfitting. 
Therefore, we divide the prompt collection into $J$ groups, with each group containing $L$ prompts.   
During training, each group of prompts is trained by the images augmented by the corresponding augmentation form, as shown in Fig.~\ref{fig:aug}. In other words, each selected augmentation operation is responsible for generating a specific type of augmented images, as well as training the corresponding group of prompts. In this way, the prompts become more diverse and can better exploit the knowledge learned in the CLIP. The probability of predicting the image can then be computed as:

\begin{equation}
    p(y_i|\mathbf{x}_{i,j})=\frac{ e^{\langle\mathbf{z}_{i,j},~\sum_{l}g(\mathbf{t}_{y_i}(\mathbf{P}_{l,j}))/L \rangle/\tau}}{\sum_{k=1}^K e^{\langle\mathbf{z}_{i,j},~\sum_{l} g(\mathbf{t}_k(\mathbf{P}_{l,j}))/L\rangle/\tau}},
    \label{eq:aug_predi}
\end{equation}
where $\mathbf{P}_{l,j}$ denotes the $l$-th prompt in the $j$-th prompt group. In this work, we set $L=8$ and $J=4$, so the total number of the prompts in the whole collection is $J\times L=32$.

\subsection{Selective Attack}
\label{sec:3.3}

\begin{figure}[t]
    \begin{center}
    \includegraphics[width=1\linewidth]{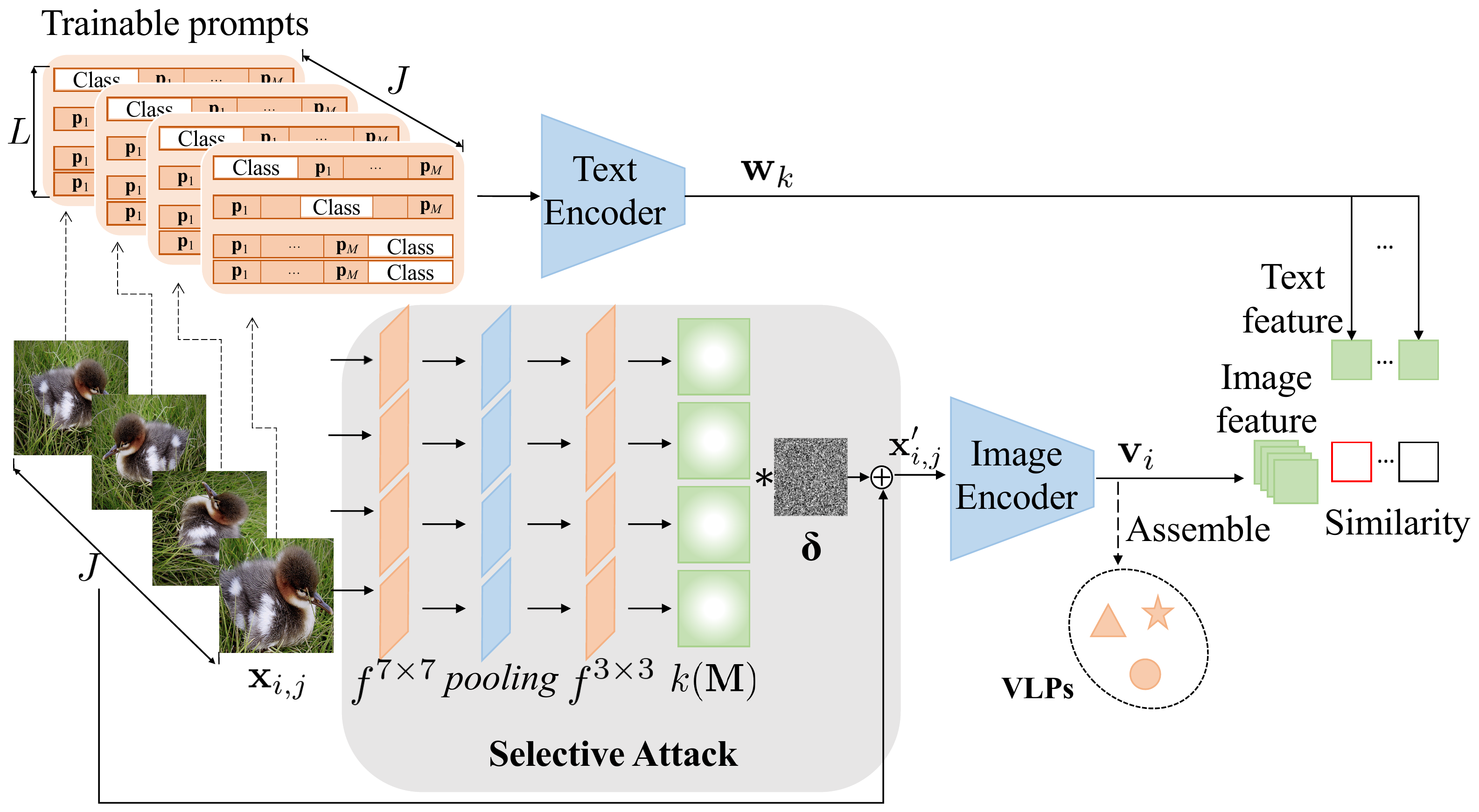}
    \caption{Framework of the proposed augmentation strategy and the Selective Attack module. The augmentation strategy associates each augmented image with a specific group of prompts to increase the diversity of the learned prompts. For each augmented image, a separate adapter and a spatial attention map are learned.}
    \vspace{-6mm}
    \label{fig:aug}
\end{center}
\end{figure}

We design a Selective Attack (SA) module, which attacks the class-irrelevant image contents to alleviate overfitting. The class-irrelevant information, such as image backgrounds, results in intra-class difference and {a distribution shift.
By attacking the class-irrelevant features, the distribution can be calibrated to better generalize to unseen samples within the same class.}

The SA module is added in front of the pre-trained image encoder as shown in Fig.~\ref{fig:aug}. The module contains two trainable adapter layers to generate a spatial attention map for each image, with the first layer as: 
\begin{equation}
    \mathbf{F}_{i,j} = \varphi( f_{j}^{7\times7}(\mathbf{x}_{i,j})),
    \label{eq:sa1}
\end{equation}
where the activation function $\varphi$ is the sigmoid, $f_j^{7\times7}$ is a convolutional layer with a kernel size of $7\times7$ operating on the $j$-th augmented image, and $\mathbf{F}_{i,j}$ is the feature obtained after the first adapter layer.

To compute the spatial attention that guides the SA on class-irrelevant areas, we then aggregate the channel information of the obtained feature $\mathbf{F}_{i,j}$ by applying channel-wise average-pooling and max-pooling, generating two 2D maps: $\mathbf{F}_{i,j}^{avg} \in \mathbb{R}^{H \times W}$ and $\mathbf{F}_{i,j}^{max} \in \mathbb{R}^{H \times W}$, where $H \times W$ is the size of the image. Applying channel-wise pooling operations has proven to be effective in highlighting informative regions~\cite{woo2018cbam}. $\mathbf{F}_{i,j}^{avg}$ and $\mathbf{F}_{i,j}^{max}$ are further concatenated and convolved by a convolutional layer with a kernel size of $3\times3$ to produce a 2D spatial attention map. In short, the process is denoted as:
\begin{equation}
    \mathbf{M}_{i,j} = \varphi( f_j^{3\times3}([\mathbf{F}_{i,j}^{avg}, \mathbf{F}_{i,j}^{max}])),
    \label{eq:sa2}
\end{equation}
where $[\cdot, \cdot]$ denotes concatenation and $\mathbf{M}_{i,j}$ is the generated spatial attention.
The larger values in the spatial attention are considered to better represent the class-relevant features, while the smaller values denote class-irrelevant contents, \emph{e.g.}, the background. We thus adopt a kernel $k(\cdot)$ to transform the spatial attention $\mathbf{M}_{i,j}$ to $k(\mathbf{M}_{i,j})=1-\mathbf{M}_{i,j}\circ\mathbf{M}_{i,j}$, thereby guiding the perturbation $\bm{\delta}$ to selectively attack the class-irrelevant regions. {
We adopt the Gaussian perturbation instead of the adversarial perturbation (\emph{e.g.}, FGSM~\cite{goodfellow2014explaining}) as the attack, since we experimentally find that the former leads to almost the same results, with significantly reduced training time.
} The attacked input is then obtained as:
\begin{equation}
    \begin{aligned}
    \mathbf{x'}_{i,j} &= \mathbf{x}_{i,j} + k(\mathbf{M}_{i,j}) \circ \bm{\delta} \\
    &= \mathbf{x}_{i,j} + (1-\mathbf{M}_{i,j} \circ \mathbf{M}_{i,j}) \circ \bm{\delta},\\
    \end{aligned}
    \label{eq:att}
\end{equation}
where $\circ$ denotes the Hadamard product and $\bm{\delta} \in \mathbb{R}^{H \times W}$. During inference, the Gaussian perturbation is no longer added, while the four groups of adapter layers (corresponding to the four types of augmented images) are averaged to obtain one group of the two adapter layers (see Fig.~\ref{fig:frame}). The attention map generated is used for calibrating the feature of the test image.

\subsection{Cross-Modal Distribution Alignment}
\label{sec:3.4}
CLIP only linearly projects the image features and the text features to the same space, whereas there exists a gap between the distributions of the image and text representations of the same class~\cite{ramesh2022hierarchical}. In order to better align the cross-modal distributions, we propose a Vision-Language Prototype (VLP) for each class to calibrate the image class prediction during inference. Specifically, we define $\mathbf{VLP} \triangleq [\mathbf{v}_1, \mathbf{v}_2, ... ,\mathbf{v}_K]$, where $\mathbf{v}_k$ is the VLP of the $k$-th class.
 
First, we construct a collection of trainable parameters $\mathbf{v}\in\mathbb{R}^{N \times J \times K \times D}$ with visual information. $\mathbf{v}_{n,j}^k\in\mathbb{R}^{D}$ is an element of $\mathbf{v}$ which is initialized by $\mathbf{z}_{n,j}^{k,0}$. $\mathbf{z}_{n,j}^{k}$ denotes the image feature of the $j$-th augmentation of the $n$-th shot in the $k$-th class, and $\mathbf{z}_{n,j}^{k,0}$ is the image feature $\mathbf{z}_{n,j}^{k}$ trained after the first epoch. $\mathbf{v}_k$ is computed as:

\begin{equation}
    \mathbf{v}_k = \frac{\sum_{n,j} \mathbf{v}_{n,j}^k}{N\times J},
    \label{eq:vp}
\end{equation}
where $N$ denotes the total number of samples in each class. Note that the initialization of VLPs is only performed after the first epoch. Then, in order to align the visual information and the language information as the VLPs, we adopt the Earth Mover's Distance (EMD) as the objective function to optimize the VLPs. EMD can well serve as a metric for computing the distance between two distributions~\cite{kline2019properties}.
Let $\mathbf{LP} \triangleq  [\mathbf{w}_1, \mathbf{w}_2, ... ,\mathbf{w}_K]$ be the language prototypes of $k$ classes with $\mathbf{w}_k$ defined as:
\begin{equation}
    \mathbf{w}_k =\frac{\sum_{l,j} \mathbf{w}_{l,j}^k}{L\times J} = \frac{\sum_{l,j} g(\mathbf{t}_k(\mathbf{P}_{l,j}))}{L\times J}.
    \label{eq:lp}
\end{equation}

\begin{figure*}[ht]
    \begin{center}
    \includegraphics[width=0.85\linewidth]{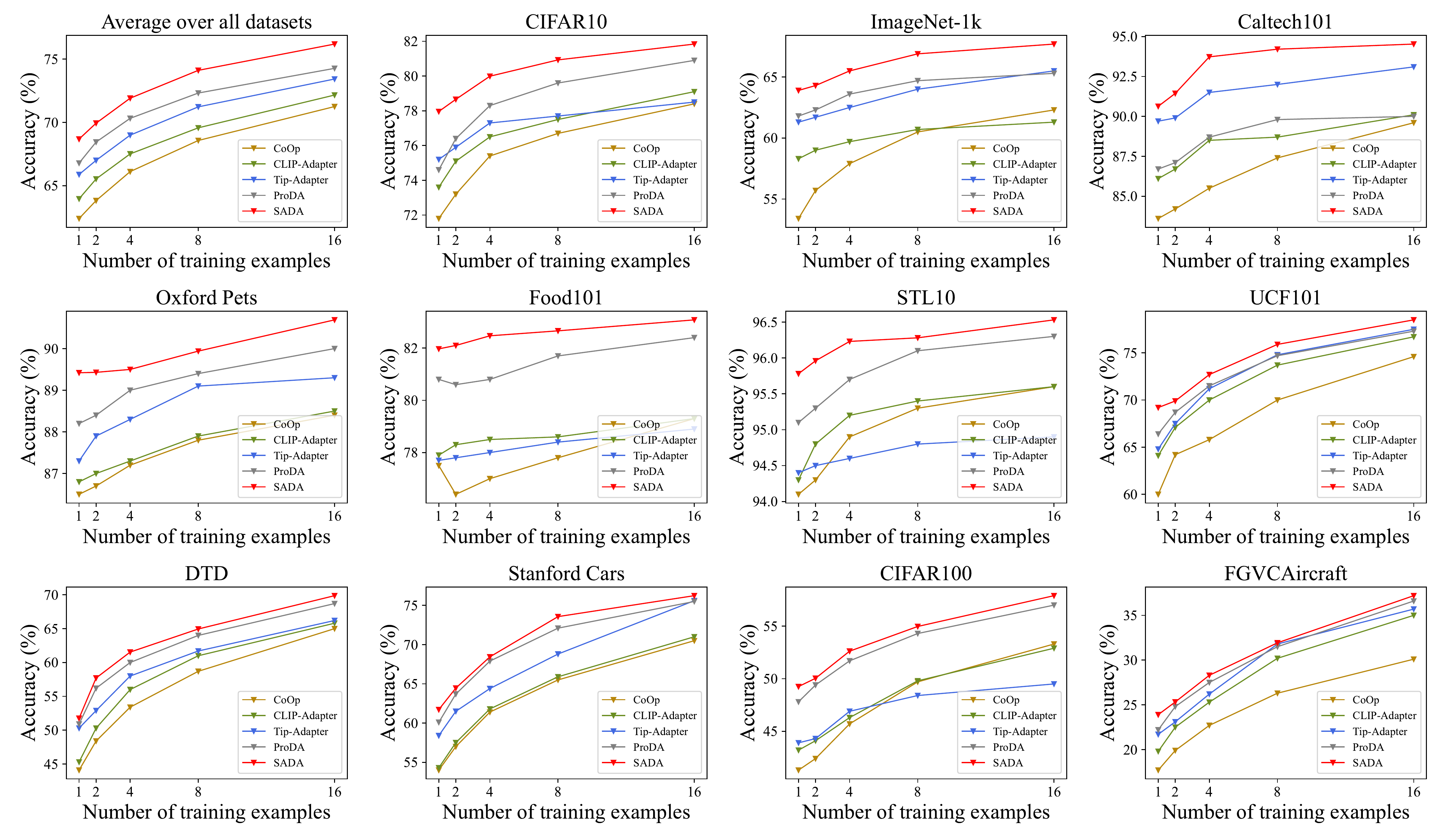}
    \caption{Main results of few-shot learning on 11 datasets. Our SADA consistently shows better performance than prior arts across different number of training samples.}
    \vspace{-6mm}
    \label{fig:result}
\end{center}
\end{figure*}

The high-level embeddings of the same class are usually adjacent, which can be modeled using a simple distribution, such as the multivariate Gaussian distribution~\cite{lu2022prompt}. Assuming that $\mathbf{v}_k\thicksim\mathcal{N}(\bm{\mu}_{\text{v}}^k,\bm{\Sigma}_\text{v}^k)$ and $\mathbf{w}_k\thicksim\mathcal{N}(\bm{\mu}_\text{w}^k,\bm{\Sigma}_\text{w}^k)$, the EMD can then be written as~\cite{clement2008elementary}:
\begin{equation}
\small
    \begin{aligned}
    \text{EMD}(\mathcal{N}(\bm{\mu}_{\text{v}}^k,\bm{\Sigma}_\text{v}^k),\mathcal{N}(\bm{\mu}_\text{w}^k,\bm{\Sigma}_\text{w}^k))
    &=\sum_k\text{EMD}(\mathbf{v}_k,\mathbf{w}_k) \\
    &=\sum_k\inf\mathbb{E}\|\mathbf{v}_k-\mathbf{w}_k\|.\\
    \end{aligned}
    \label{eq:emd}
\end{equation}
The complexity of the EMD algorithm is $\mathcal{O}(D^3\log D)$~\cite{rubner2000earth} and $D=1024$ in this work. To speed up the training, we derive an upper bound for the EMD on the multivariate Gaussian distributions, and adopt this bound as the objective function to update the VLPs. Based on Jensen's inequality~\cite{chandler1987introduction}, the upper bound of EMD is derived as:
\begin{equation}
    \mathcal{L}_{\text{EMD}}\triangleq \sum_k(\|\bm{\mu}_{\text{v}}^k-\bm{\mu}_{\text{w}}^k\|^2+\|{\bm{\Sigma}_{\text{v}}^k}^{\frac{1}{2}}-{\bm{\Sigma}_{\text{w}}^k}^{\frac{1}{2}}\|^2).
    \label{eq:upbound}
\end{equation}
The detailed derivation of the upper bound is given in the supplementary materials. The complexity of computing $\mathcal{L}_{\text{EMD}}$ now becomes $\mathcal{O}(D)$.
In addition to the alignment loss $\mathcal{L}_{\text{EMD}}$, we also need a classification loss, which is defined based on Eqs.~\ref{eq:aug_predi} and \ref{eq:vp} as:
\begin{equation}
        \mathcal{L}_m=\mathbb{E}[-\text{log}\frac{ e^{\langle(1-\alpha)\mathbf{z}_{i,j}+\alpha\mathbf{v}_{y_i},~\sum_{l} g(\mathbf{t}_{y_i}(\mathbf{P}_{l,j}))/L\rangle/\tau}}{\sum_{k=1}^K e^{\langle(1-\alpha)\mathbf{z}_{i,j}+\alpha\mathbf{v}_{y_i},~\sum_{l} g(\mathbf{t}_k(\mathbf{P}_{l,j}))/L\rangle/\tau}}],
    \label{eq:loss_main}
\end{equation}
where $\alpha \in (0,1)$ is a hyper-parameter that denotes the distribution calibration ratio\footnote{The geometric explanation of why $(1-\alpha)\mathbf{z}_{i,j}+\alpha\mathbf{v}_{y_i}$ in Eq.~\ref{eq:loss_main} helps the alignment is given in the supplementary materials.}, $y_i$ is the class label for $\mathbf{x}_i$, and $\mathbf{v}_{y_i}$ is the VLP of the class $y_i$. During training, the VLPs are updated by $\mathcal{L}_{\text{EMD}}$ and $\mathcal{L}_m$, while the adapter layers and the prompts are updated by $\mathcal{L}_m$. During inference, the labels of the test images are unavailable. Therefore, we adopt the VLPs to calibrate the image predictions by calculating a normalized weighting vector $\mathbf{\bar{d}}$ of $\mathbf{v}_k$ as: 
\begin{equation}
\small
    \mathbf{d}=(d_1, d_2, \dots, d_K)^T, \ d_k=\frac{1}{\| \mathbf{z}_{i}- \mathbf{v}_k\|}, \ k=1, 2, \dots, K,
    \label{eq:d}
\end{equation}
\begin{equation}
\small
    \mathbf{\bar{d}}=(\bar{d_1}, \bar{d_2}, \dots, \bar{d_K})^T, \ \bar{d_k}=\frac{d_k}{\sum_{m=1}^{K}{d_m}}, \ k=1, 2, \dots, K,
    \label{eq:d_norm}
\end{equation}
Then, the probability of predicting the image after the cross-modal distribution alignment is computed as:
\begin{equation}
\small
    p(y_i|\mathbf{x}_{i})=\frac{ e^{\langle(1-\alpha)\mathbf{z}_i+\alpha (\mathbf{\bar{d}}^T\mathbf{VLP})^T,~\sum_{l} g(\mathbf{t}_{y_i}(\mathbf{P}_{l,j}))/L \rangle/\tau}}{\sum_{k=1}^K e^{\langle(1-\alpha)\mathbf{z}_i+\alpha (\mathbf{\bar{d}}^T\mathbf{VLP})^T,~\sum_{l} g(\mathbf{t}_k(\mathbf{P}_{l,j}))/L\rangle/\tau}}.
    \label{eq:predi_final}
\end{equation}
$p(y_i|\mathbf{x}_{i})$ is finally used to predict the classes of the test image samples. 

\vspace{-1mm}
\section{Experiments}
\label{sec.4}
\vspace{-1mm}
In this section, we first compare our method (termed SADA) with prior arts on 11 datasets, and show that SADA achieves best results on all the datasets. Then, the specific effect of each proposed module is analyzed. We implement our model using the MindSpore Lite tool~\cite{mindspore}.
\vspace{-1mm}
\subsection{Implementation Details}
\label{sec:4.1}
\vspace{-1mm}
\textbf{Datasets.} 
The 11 classification datasets cover a diverse set of benchmarks including CIFAR10~\cite{krizhevsky2009learning}, ImageNet-1k~\cite{deng2009imagenet}, Caltech-101~\cite{fei2004learning}, Oxford-IIIT Pets~\cite{parkhi2012cats}, Food-101~\cite{bossard2014food}, STL-10~\cite{coates2011analysis}, UCF-101~\cite{soomro2012ucf101}, DTD~\cite{cimpoi2014describing}, Stanford Cars~\cite{krause20133d}, CIFAR100~\cite{krizhevsky2009learning} and FGVC Aircraft~\cite{maji2013fine}.
Our experiments follow the few-shot training and evaluation protocol of CLIP, in which 1, 2, 4, 8, and 16 labeled images per class on each dataset are randomly sampled for training. The average evaluation results over 10 runs are presented.

\textbf{Baselines.} We compare our SADA with the most related and recent models CoOp~\cite{zhou2022learning}), CLIP-Adapter~\cite{gao2021clip},  Tip-Adapter~\cite{zhang2021tip}, and ProDA~\cite{lu2022prompt}).
The results of linear-probe CLIP are much worse than those of these methods, and are only given in the supplementary materials.

\textbf{Training Details.} 
For a fair comparison, we adopt CLIP's ResNet-50 as our image encoder and CLIP's Transformer as our text encoder, which are also used in ProDA, CoOp and CLIP-Adapter.
The prompt length $M$ is set to 16, and the total number of prompts in the collection is 32. The distribution calibration ratio $\alpha$ is 0.1. 
We train the model for 50 epochs using SGD with an initial learning rate of 0.001 for $\mathcal{L}_m$ and 0.01 for $\mathcal{L}_{\text{EMD}}$, both following a cosine decay schedule. The prompt batch size is 4, and the image batch size is 20. {
The Gaussian perturbation is sampled from $\mathcal{N}(0, 0.7^2)$.
} The model of the last training epoch is used for evaluation.
\vspace{-1mm}
\subsection{Main Results}
\vspace{-1mm}
Fig.~\ref{fig:result} shows the comparison results on the 11 datasets. The average results by the models over all the datasets are also provided in the first sub-figure of Fig.~\ref{fig:result}. Our SADA significantly outperforms the baselines and achieves best results under all the shot numbers. This demonstrates the generalization ability of SADA to learn quickly from a small number of samples. The specific values of the curves are given in the supplementary materials.

Compared with the previous best model ProDA~\cite{lu2022prompt}, our SADA consistently outperforms it on the average results. For example, SADA improves the results of ProDA by 1.90\% and 1.92\% under 1-shot and 16-shot settings, respectively. On some specific datasets, our SADA achieves more significant improvements. For example, SADA improves ProDA by 3.36\%, 2.80\% and 2.10\% under 1-shot on CIFAR10, UCF-101 and ImageNet-1k, respectively. On more challenging fine-grained datasets such as Food-101, Oxford-IIIT Pets, Stanford Cars, and FGVC Aircraft, our method still achieves better results. 

\vspace{-1mm}
\subsection{Ablation Study}
\label{sec:4.2}
\begin{figure}[t]
    \centering
    \includegraphics[width=0.85\linewidth]{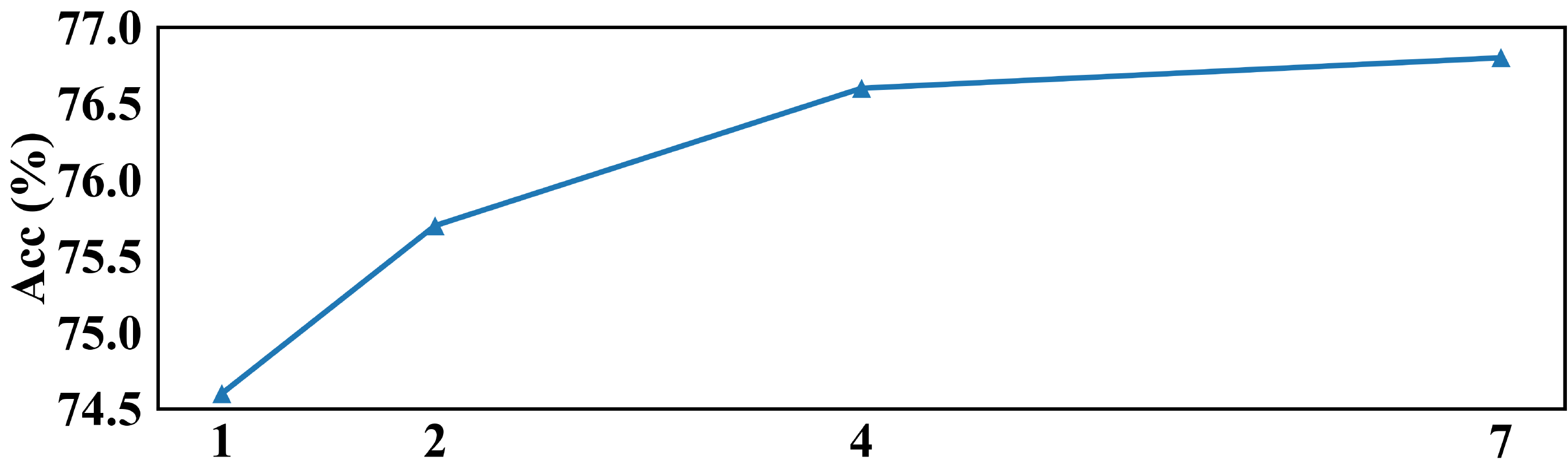}
    \vspace{-3mm}
    \caption{Test accuracy ($\%$) of training with different numbers of augmentation operations on CIFAR10.}
    \label{fig:result_aug}
    \vspace{-2mm}
\end{figure}

\vspace{-1mm}
\textbf{Different numbers of
augmentation operation.}
As introduced in Sec.~\ref{sec:3.2}, we propose an augmentation strategy to mitigate overfitting and increase the diversity of the images and the text prompts. We first evaluate the effect of the number of the augmentation operations on the test results.
The candidate pool of augmentation operations consists of \textit{rotating, flipping, random cropping$+$resizing, random gray scaling, resizing, color jittering}, and \textit{Gaussian blurring}. We compare four cases where the operation number $J$ is set to $1,2,4,7$, respectively. When $J=1$, the augmentation with the best test results is \textit{flipping}. When $J=2$, the best operations are \textit{flipping} and \textit{random gray scaling}. When $J=4$, the best operations are \textit{flipping}, \textit{Gaussian blurring}, \textit{random gray scaling}, and \textit{random cropping$+$resizing}. When $J=7$, all the operations are adopted. The performances of these cases are shown in Fig.~\ref{fig:result_aug}. Considering the trade-off between the performance and the computation consumption, we choose $J=4$ in our experiments.

\textbf{Prompt diversity.} We further verify the effect of the augmentation on the prompt diversity. The 32 prompts in the collection are divided into 4 augmentation groups ($J=4$) as shown in Fig.~\ref{fig:aug}. Let SADA w/o Aug be the SADA model but without the data augmentation. In Table~\ref{table1}, the mean values of all the prompts in each group obtained by SADA w/o Aug and SADA are given. Then we calculate the standard deviation (std) of these 4 mean values of each model. The std of SADA is significantly larger than that of SADA w/o Aug, demonstrating larger prompt diversity after the data augmentation.

\begin{table}[t]\centering
\caption{Effect of the augmentation on prompt diversity.
}
\vspace{-3mm}
\resizebox{1\columnwidth}{!}{
\begin{tabular}{c|cccc|c}
\toprule
\multirow{2}{*}{Group j} & \multicolumn{4}{c|}{Mean}         & \multirow{2}{*}{Std} \\ \cline{2-5}
                         & 1      & 2      & 3      & 4      &                      \\ \toprule
SADA w/o Aug             & 0.0954 & 0.0923 & 0.0892 & 0.0998 & 0.0045               \\
SADA                     & 0.1035 & 0.1441 & 0.0855 & 0.1173 & \textbf{0.0247}               \\ \bottomrule
\end{tabular}}
\vspace{-2mm}
\label{table1}
\end{table}

\begin{table}[t]\centering
\caption{Ablation of SA and CMDA on CIFAR10.}
\vspace{-3mm}
\resizebox{1\columnwidth}{!}{
\begin{tabular}{c|ccccc}
\toprule
\#Shots                   & 1 & 2 & 4 & 8 & 16 \\ \toprule
Baseline                  &74.61\%   &76.40\%   & 78.34\%  & 79.63\%  & 80.90\%   \\
Baseline w SA             & 77.61\%  & 78.2\%  & 79.63\%  & 80.53\%  & 81.38\%   \\
Baseline w CMDA & 76.79\%  & 77.37\%  &79.02\%   & 80.15\%  & 81.31\%   \\
 \bottomrule
\end{tabular}}
\vspace{-2mm}
\label{table2}
\end{table}


\textbf{Ablation of SA and CMDA.}
In this section, we conduct ablation studies on CIFAR10. First of all, we define three models for evaluation: {1) Baseline, in which we remove the SA and CMDA modules, and replace $(1-\alpha)\mathbf{z}_{i,j}+\alpha\mathbf{v}_{y_i}$ in Eq.~\ref{eq:loss_main} and $(1-\alpha)\mathbf{z}_i+\alpha (\mathbf{\bar{d}}^T\mathbf{VLP})^T$ in Eq.~\ref{eq:predi_final} with $\mathbf{z}_{i,j}$ and $\mathbf{z}_i$, respectively; 2) Baseline w SA, in which we add the SA module to Baseline; 3) Baseline w CMDA, in which we add the CMDA module to Baseline during both training and inference.} 
In particular, the 1-shot case shows 3\% (74.61\% vs. 77.61\%) and 2.18\% (74.61\% vs. 76.79\%) improvements by Baseline w SA and Baseline w CMDA, respectively. Combining all the modules, the full SADA gets the best results in all cases.


\begin{figure}[t]
    \centering
    \includegraphics[width=0.85\linewidth]{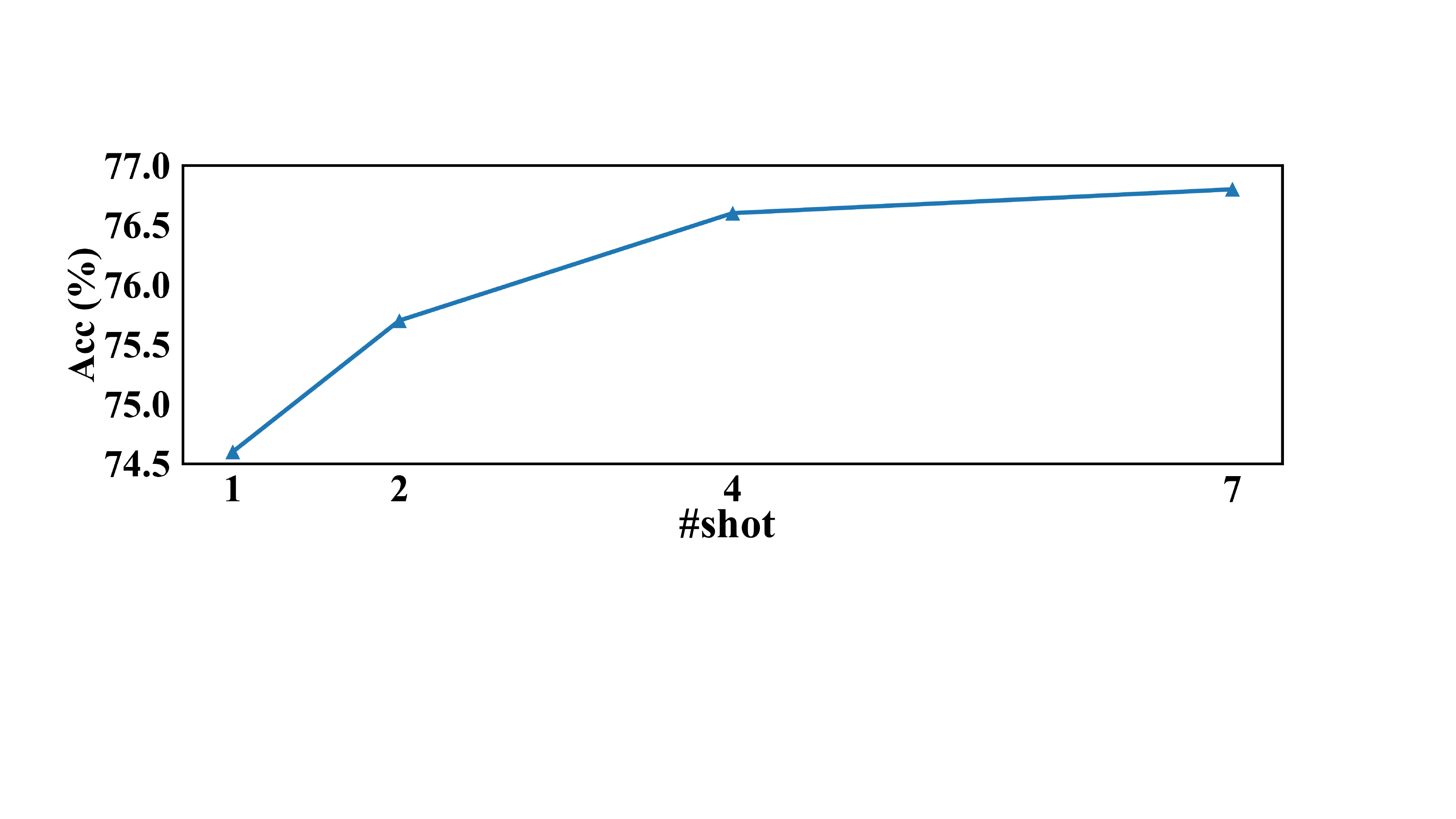}
    \vspace{-2mm}
    \caption{1-shot accuracy ($\%$) of different attack strength.}
    \label{fig:attack}
    \vspace{-5mm}
\end{figure}

{\textbf{Attack strength of SA.}
In the SA module, the Gaussian perturbations are sampled from $\mathcal{N}(0,\sigma^2)$. We further train the model by varying $\sigma$ from 0 to 0.9, and report the testing accuracies on CIFAR10 in Fig.~\ref{fig:attack}, where $\sigma\!=\!0$ means naively adding two trainable layers before the pre-trained image encoder without imposing any attack on the image. Compared with no attack ($\sigma\!=\!0$), introducing Gaussian perturbations significantly improves the testing accuracy. This demonstrates that SA improves performance not only because it introduces new trainable parameters, but also because the attack plays its role in removing image redundancy. We set $\sigma\!=\!0.7$ (where the performance is optimal) for all the other experiments.
}

\begin{figure}[t]
    \centering
    \includegraphics[width=0.85\linewidth]{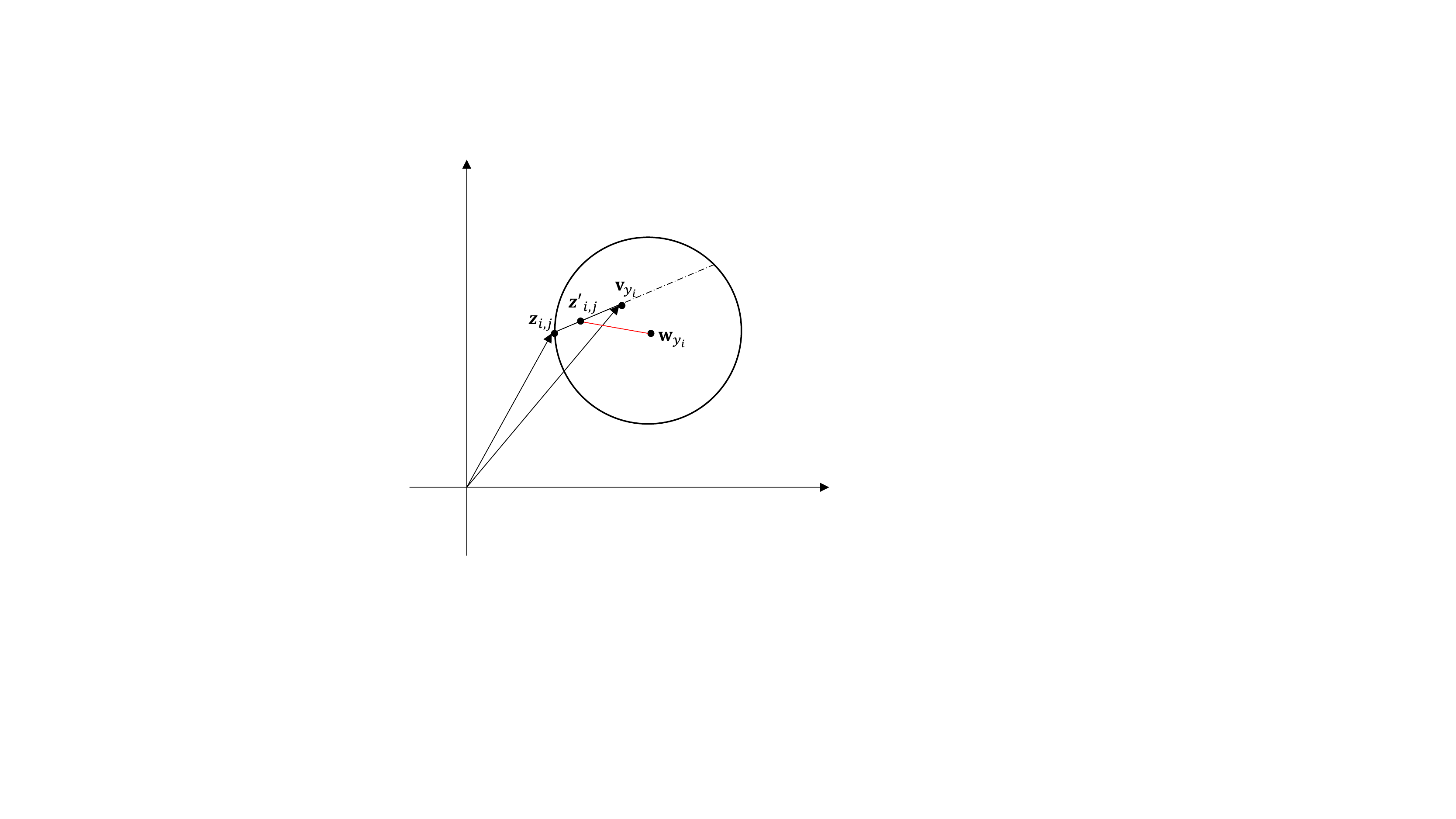}
    \vspace{-3mm}
    \caption{{1-shot accuracy ($\%$) on CIFAR10 when SA is at different layers of the image encoder.}}
    \label{fig:position}
    \vspace{-2mm}
\end{figure}

\begin{figure}[t]
    \centering
    \includegraphics[width=0.85\linewidth]{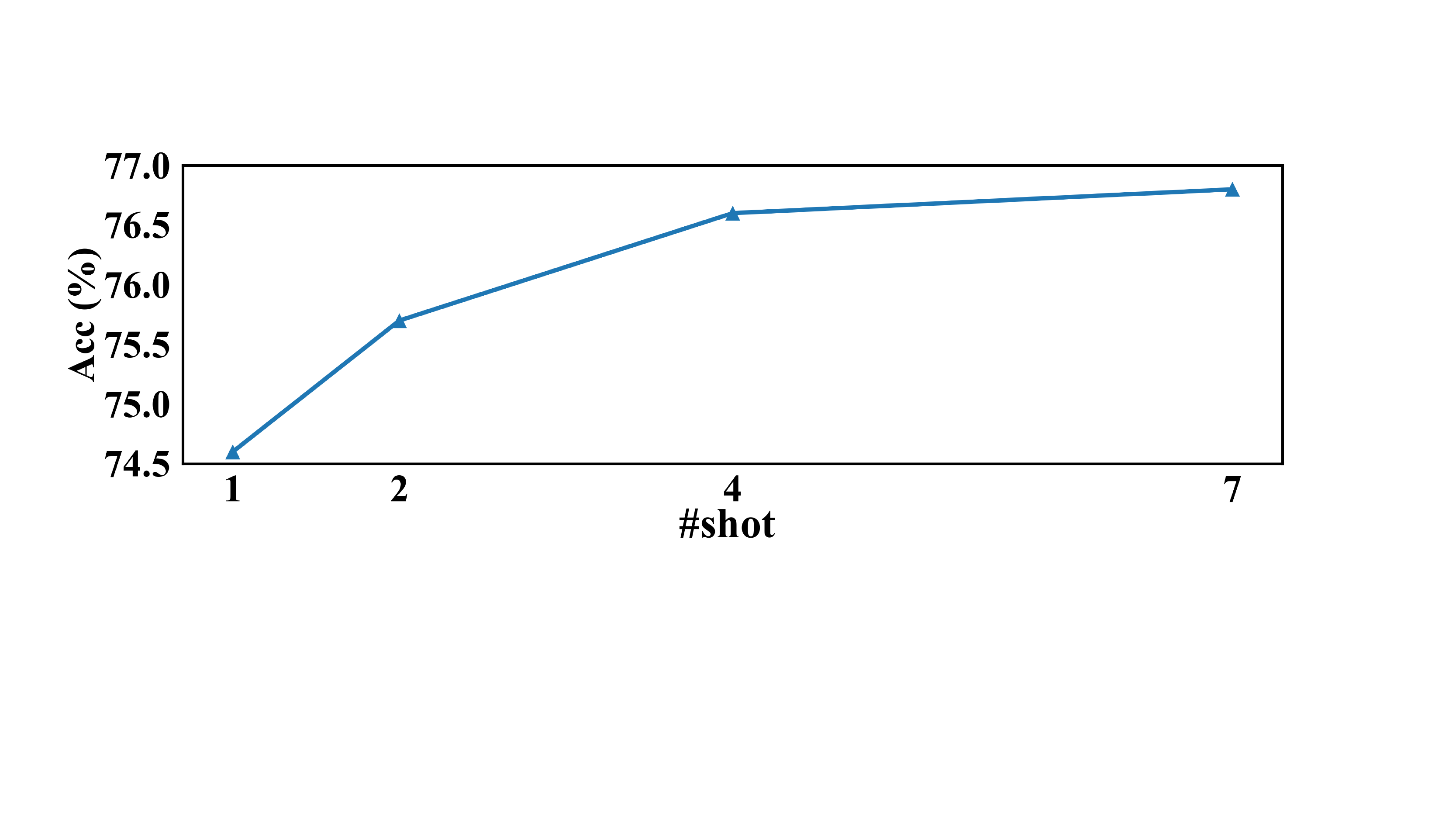}
    \vspace{-3mm}
    \caption{1-shot accuracy ($\%$) of different calibration ratio $\alpha$.}
    \label{fig:alpha}
    \vspace{-2mm}
\end{figure}

{\textbf{Position of SA module.}
We further evaluate which layer to attach the SA module to. We place the SA module at the input layer (as in Fig.~\ref{fig:aug}), or after the first, second, third or fourth block of ResNet-50. Fig.~\ref{fig:position} shows that the performance suffers from significant degradation when the module is placed inside instead of in front of the encoder. We intuitively owe this result to the facts that 1) placing trainable layers inside the encoder destroys the prior stored in the pre-trained weights, and 2) adding perturbations to higher-level features of deeper layers affects the classification results more seriously.
}


\textbf{Calibration Ratio $\alpha$.}
We test different distribution calibration ratio $\alpha$ on CIFAR10. As shown in Fig.~\ref{fig:alpha}, the performance is the best when $\alpha = 0.1$. On other datasets, we also have this similar phenomenon, so we choose $\alpha = 0.1$ in all the experiments.

\begin{table}[t]\centering
\small
\caption{{Ablation on the objective function to optimize the VLPs.
}}
\vspace{-2mm}
\resizebox{1\columnwidth}{!}{
\begin{tabular}{c|ccccc}
\toprule
\#Shots     & 1 & 2 & 4 & 8 & 16 \\ \toprule
EMD    &\textbf{76.7}\%	&\textbf{77.3}\%	&\textbf{79.0}\%	&\textbf{80.1}\%	&\textbf{81.3}\%   \\	
MMD    & 73.5\%	  &76.1\%	 &77.4\%	 &79.7\%	  &80.5\%   \\	
JS-Divergence  & 74.3\%  & 75.9\%  &77.6\%   &79.2\%  & 80.1\%   \\

\bottomrule
\end{tabular}}
\vspace{-5mm}
\label{table3}
\end{table}

{\textbf{EMD.}  In Table~\ref{table3}, we verify that the Earth Mover’s Distance (EMD) is an effective objective function to optimize the VLPs. We compare EMD with two other measures of distribution difference, \emph{i.e.}, MMD~\cite{gretton2012kernel} and JS-Divergence~\cite{fuglede2004jensen}. Experimental results on CIFAR10 show that EMD outperforms the other two functions in all cases of shots.}

\begin{figure}[t]
    \centering
    \includegraphics[width=0.85\linewidth]{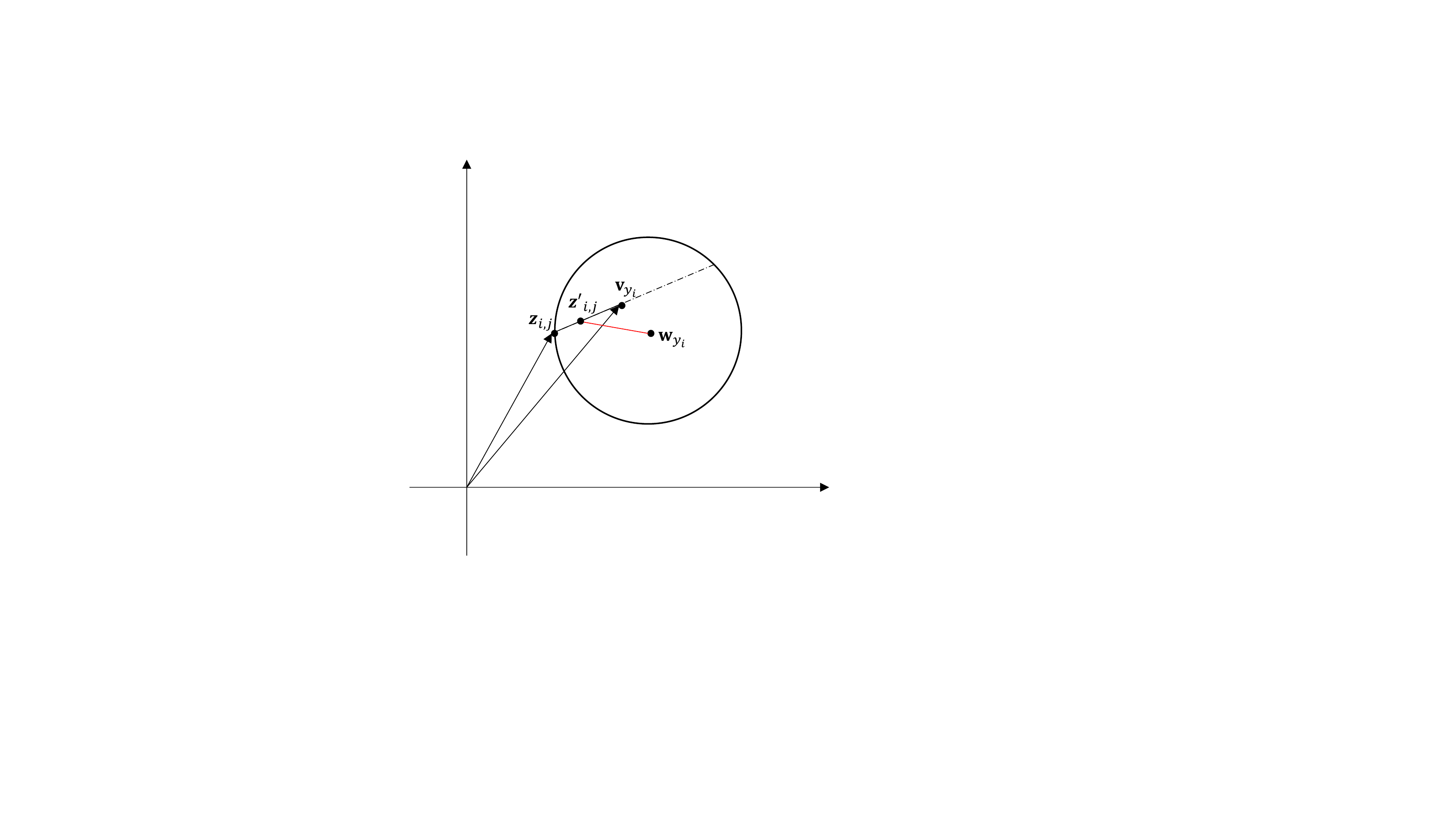}
    \vspace{-2mm}
    \caption{Effect of VLPs on CIFAR10.}
    \label{fig:da_abla}
    \vspace{-2mm}
\end{figure}

\textbf{Effect of $\mathbf{VLP}$ in cross-modal distribution alignment.}
We verify the effect of Vision-Language Prototypes (VLPs) in Fig.~\ref{fig:da_abla} with three models. 1) Baseline is defined in Table~\ref{table2}. 2) Baseline w VLP aligns the cross-modal distribution by $\mathbf{VLP}$. 3) Baseline w LP is the same as Baseline w VLP in except that $\mathbf{v}_{y_i}$ in Eq.~\ref{eq:loss_main} and $\mathbf{VLP}$ in Eq.~\ref{eq:predi_final} are replaced  with $\mathbf{w}_{y_i}$ and $\mathbf{LP}$, respectively. Baseline w VLP delivers a performance boost in all shot cases. In particular, the 1-shot case shows a 2.18\% (74.61\% vs. 76.79\%) improvement over Baseline.

\begin{figure}[t]
    \centering
    \includegraphics[width=0.65\linewidth]{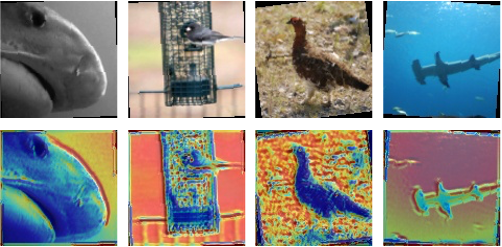}
    \vspace{-2mm}
    \caption{Visualization of attacked areas (in red) guided by $1-\mathbf{M}\circ\mathbf{M}$. The images are from ImageNet-1k.}
    \label{fig:mask}
    \vspace{-5mm}
\end{figure}

\subsection{Visualization of Selective Attack and CMDA}
\label{sec:4.6}

The Selective Attack module attacks the class-irrelevant information of the images, reduces the intra-class distances of image features, and helps to avoid overfitting.
We visualize the kernelized spatial attention in Fig.~\ref{fig:mask}, in which the red areas denote higher attention values, while the blue areas denote lower attention values. We can see that mainly the background areas are given higher attention weights to guide the selective attack.

As shown in Figs.~\ref{fig:motivation} (a) and (b), after Selective Attack, the intra-class image representations become more clustered as expected.
We also verify the alignment effect of CMDA in Figs.~\ref{fig:motivation} (c) and (d), the difference between the two distributions is significantly reduced.

\vspace{-2mm}
\section{Conclusion}
\vspace{-1mm}
This paper proposes a few-shot learning method with visual distribution calibration and cross-modal distribution alignment (CMDA) based on a pre-trained vision-language model. The Selective Attack module eliminates class-irrelevant information in the images and calibrate the visual distribution. The CMDA aligns the distributions of the image features and the text features. Overall, we improve the performance of the few-shot learning and achieve state-of-the-art results on 11 datasets. In future work, we will explore the potential of our method in other applications.

\vspace{-2mm}
\section*{Acknowledgements}
\vspace{-1mm}
This work was supported by National Natural Science Foundation of China under Grant 62076016 and 62141604, Beijing Natural Science Foundation L223024. We gratefully acknowledge the support of MindSpore \cite{mindspore}, CANN (Compute Architecture for Neural Networks) and Ascend AI Processor used for this research. 

{\small
\bibliographystyle{ieee_fullname}
\bibliography{egbib}
}

\end{document}